\documentclass[lettersize,journal]{IEEEtran}
\usepackage{amsmath,amsfonts}
\usepackage{algorithmic}
\usepackage{algorithm}
\usepackage{array}
\usepackage[caption=false,font=normalsize,labelfont=sf,textfont=sf]{subfig}
\usepackage{textcomp}
\usepackage{stfloats}
\usepackage{url}
\usepackage{verbatim}
\usepackage{graphicx}
\usepackage{cite}
\usepackage{bm}
\usepackage{multirow}
\usepackage{multicol}
\usepackage{booktabs}
\usepackage{caption}
\usepackage{amsfonts}
\usepackage{amsmath}
\usepackage{colortbl}
\usepackage{xcolor}
\usepackage{amssymb}
\usepackage{mathtools}

\def\etal{\emph{et al.}}

\begin{document}

\title{Anisotropic Pooling for LUT-realizable CNN Image Restoration}

\author{Xi~Zhang,~\IEEEmembership{Member,~IEEE}, and
        Xiaolin~Wu,~\IEEEmembership{Life Fellow,~IEEE}
\thanks{X.~Zhang is with the ANGEL Lab, Nanyang Technological University, Singapore. (email: xi.zhang@ntu.edu.sg).}
\thanks{X.~Wu is with the School of Computing and Artificial Intelligence, Southwest Jiaotong University, Chengdu, China (email: xwu510@gmail.com).}
}

\markboth{Journal of \LaTeX\ Class Files,~Vol.~14, No.~8, August~2021}%
{Shell \MakeLowercase{\textit{et al.}}: A Sample Article Using IEEEtran.cls for IEEE Journals}


\maketitle

\begin{abstract}
Table look-up realization of image restoration CNNs has the potential of achieving competitive image quality while being much faster and resource frugal than the straightforward CNN implementation.  The main technical challenge facing the LUT-based CNN algorithm designers is to manage the table size without overly restricting the receptive field.  The prevailing strategy is to reuse the table for small pixel patches of different orientations (apparently assuming a degree of isotropy) and then fuse the look-up results.  The fusion is currently done by average pooling, which we find being ill suited to anisotropic signal structures.  
To alleviate the problem, we investigate and discuss anisotropic pooling methods to replace naive averaging for improving the performance of the current LUT-realizable CNN restoration methods.
First, we introduce the method of generalized median pooling which leads to measurable gains over average pooling. We then extend this idea by learning data-dependent pooling coefficients for each orientation, so that they can adaptively weigh the contributions of differently oriented pixel patches. Experimental results on various restoration benchmarks show that our anisotropic pooling strategy yields both perceptually and numerically superior results compared to existing LUT-realizable CNN methods.

\end{abstract}

\begin{IEEEkeywords}
Anisotropic pooling, Look-up table, image restoration, CNN inference.
\end{IEEEkeywords}

\section{Introduction}
\label{sec:intro}

\IEEEPARstart{R}{ecently} there has been an increased push toward efficient and hardware-friendly implementations of image restoration neural networks~\cite{SRLUT,MULUT,SPLUT,RCLUT,DFCLUT,li2025tinylut}. Conventional deep learning based methods, such as convolutional neural networks or transformer-based alternatives, can perform impressively on super-resolution, denoising, and other restoration prob
~\cite{dong2015compression,zhang2017beyond,zhang2018ffdnet,wang2018esrgan,zhang2020deblurring,agdl,davd,mdvd,zhang2023lvqac,zhang2024learning,shi2025vmambair,xu2025multirate,zeng2025low,zhang2025dc,vien2025gradient}. 
However, they typically demand large numbers of floating-point operations and substantial memory footprints, making them ill-suited for resource-constrained applications. By contrast, \emph{look-up table (LUT)} techniques have proven to be an effective lightweight alternative, in which the bulk of the computational burden is offloaded to a precomputed mapping from degraded input patches to high-quality (restored) outputs. Once stored, these mappings facilitate near-instant retrieval of restored patches at inference time.

\begin{figure}[t]
    \centering
    \includegraphics[width=\linewidth]{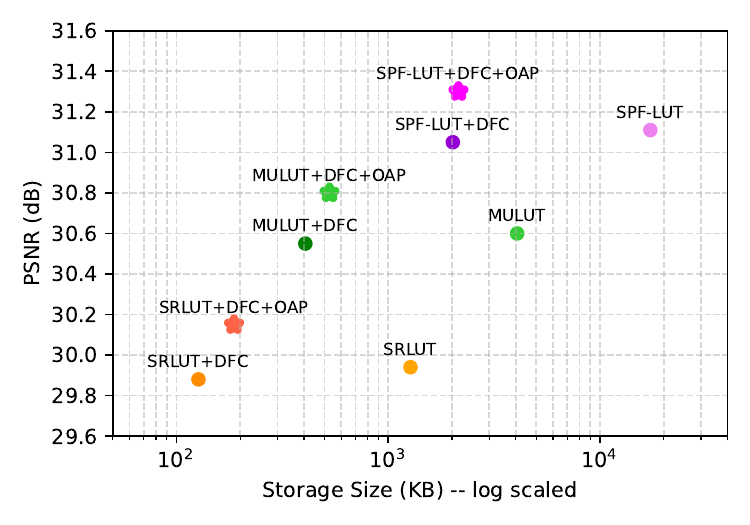}
    \caption{
    Trade-off between PSNR and storage size for the $\times4$ image super-resolution task. 
    The proposed orientation-aware pooling (OAP) mechanism consistently improves the reconstruction quality of LUT-based super-resolution models, while introducing only a negligible increase in storage. 
    }
    \label{fig:highlight}
\end{figure}

In many of the current state-of-the-art LUT-based methods, \emph{cascaded LUTs}~\cite{MULUT,SPLUT,RCLUT,DFCLUT} are employed to refine image quality progressively. Each LUT in the cascade handles a portion of the restoration process, incrementally improving the visual fidelity of the output. While the approach significantly boosts reconstruction accuracy, a common trade-off in LUT-based pipelines is the use of a relatively small receptive field around each degraded pixel or patch. Because the LUT size grows exponentially with the dimension of the input pixel patch, these patches have to be small (e.g., 2$\times$2) for the sake of practicality. To work around this limitation, the LUT methods use a multi-orientation pooling strategy (or called rotation ensemble)~\cite{SRLUT,MULUT,DFCLUT}, rotating a 3$\times$3 patch in different orientations so that an effective 3$\times$3 region is covered by a $2\times2$ convolution kernel when the rotated outputs are merged. This multi-orientation pooling allows the system to gather information from slightly beyond the local 2$\times$2 neighborhood, enhancing its ability to capture textures and edges.

Despite the cleverness of the above rotation technique, the final pooling of rotated predictions currently is averaging them. Averaging treats all orientations equally.  Such an isotropic treatment fails to account for strongly directional features such as edges and textures, which may lead to blurred textures or broken edges when strong orientation patterns are present.
To alleviate this problem, we investigate and discuss anisotropic pooling methods that can improve the current LUT-realizable CNN restoration methods. We start by analyzing the drawbacks of average pooling and then introduce two ways to improve it. 
First, we present a generalized median pooling approach (GMP) that mitigates the effects of outliers by relying on values closer to the median of the four look-up results in the four orientations. 
We then extend this idea by learning data-dependent pooling coefficients for each orientation, so that they can adaptively weigh the contributions of differently oriented pixel patches for the best restoration.  The second method is called orientation-aware pooling (OAP).
Unlike static averaging, these learned QAP coefficients generate the restored pixel as an adaptively weighted sum 
of different directional estimates.

Implementing more sophisticated pooling logic, however, raises concerns about computation and memory overhead---especially for multi-stage LUT-based pipelines, for which efficiency is paramount. To address this, we design our OAP mechanism as a small plug-in LUT that predicts orientation-pooling coefficients from the input patch. This \emph{pooling LUT} is jointly optimized with the main restoration LUTs, yet remains small enough to avoid ballooning the overall parameter count. With this pooling LUT plugged into each stage of the cascaded pipeline, the restoration LUTs in successive stages share the same adaptive pooling weights.  In other words, the small plug-in LUT is accessed only once, the directional weights can be reused at every stage. As such, the extra overhead incurred by OAP is negligible relative to the multi-LUT restoration architecture in the current literature.

In summary, this paper makes the following contributions to LUT-realizable CNN image restoration:
\begin{itemize}
    \item We thoroughly analyze the shortcomings of simple average pooling in LUT-realizable image restoration, identifying issues such as blurring or loss of sharp edges due to uniform weighting across orientations.
    \item We propose the generalized median pooling (GMP) as a robust alternative to average pooling, mitigating the influence of outlier directional estimates.
    \item We develop an orientation-aware pooling (OAP) mechanism to optimize weights of different directional estimates, favoring the more important orientations.
    \item Through extensive experiments on standard image restoration benchmarks, we show that both GMP and OAP consistently produce sharper details and higher quantitative metrics, with minimal overhead compared to existing LUT-realizable CNN approaches.
\end{itemize}

\section{Related Work}
\label{sec:related}

The \emph{Look-Up Table (LUT)} operator stands out in image processing~\cite{Mantiuk2008,de2008reducing, Lefkimmiatis2009,lee2010new,Pouli2011,Rashid2011} due to its ability to rapidly handle data through simple index-based queries. By recording index-value tuples in a multi-dimensional matrix, LUTs minimize computational effort via straightforward coordinate lookups, making them exceptionally effective in scenarios where fast data retrieval is critical.

Jo \textit{et al.} have pushed this concept further with SR-LUT~\cite{SRLUT}, a highly efficient method for image super-resolution. Their technique starts by training a deep super-resolution network with a constrained receptive field (RF) and then caching the resulting outputs in a LUT. At test time, high-resolution (HR) image predictions can be obtained by direct lookups on the low-resolution (LR) input patches. Unlike the standard color-to-color three-dimensional LUTs used in image enhancement~\cite{zeng2020learning,zhang2022clut}, SR-LUT relies on a patch-to-patch mapping in four dimensions, yielding HR details corresponding to each LR patch. Despite its speed advantages, SR-LUT faces a challenge: when the receptive field grows larger, the LUT can explode in memory requirement. 

Subsequently, MuLUT~\cite{MULUT,li2024toward} and SPLUT~\cite{SPLUT} proposed different strategies to mitigate this memory burden. SPLUT uses a cascade of additional LUTs, but its indexing requires multiple LUT stages to broaden the RF even by small increments. On the other hand, MuLUT broadens the receptive field more effectively using index schemes that complement each other, thereby optimizing the trade-off between memory and coverage.

Recent developments have introduced RCLUT~\cite{RCLUT}, which harnesses a \emph{reconstructed convolution} (RC) block. By segmenting the spatial and channel-wise computations, RCLUT expands its receptive field while keeping storage overhead relatively low. DFC-LUT~\cite{DFCLUT}, applies a \emph{diagonal-first compression} (DFC) scheme. This method prioritizes diagonal HQ/LQ entries and remaps them to conserve representational fidelity, while non-diagonal data is selectively sampled. The end result is an effective compromise between output quality and memory constraints. Another recent LUT-based approach, TinyLUT~\cite{li2025tinylut}, utilizes a novel separable mapping strategy that converts the LUT’s storage cost from exponential to effectively linear in the kernel size, alongside a dynamic discretization mechanism for further compression, thereby achieving state-of-the-art accuracy and speed for image restoration on edge devices.

In addition to these, several very recent works have further expanded LUT-based methodologies. \textit{Xu et al.} introduced AutoLUT~\cite{AutoLUT}, which addresses the inflexibility of manual sampling patterns by learning an \emph{Automatic Sampling} (AutoSample) strategy alongside an \emph{Adaptive Residual Learning} (AdaRL) module. By allowing the network to automatically select pixel samples and by reintroducing residual connections (previously avoided due to LUT value range issues), AutoLUT significantly expands the effective receptive field and improves fine detail reconstruction, all without increasing inference cost. This plug-and-play approach yields notable PSNR gains when integrated into frameworks like MuLUT and DFC-LUT, demonstrating the benefit of adaptive sampling and better feature fusion in LUT-based super-resolution.

Meanwhile, \textit{Yang et al.} proposed DnLUT~\cite{DnLUT}, a LUT-driven framework tailored for image denoising. DnLUT introduces a pairwise channel mixing module to capture cross-channel correlations and a novel \emph{L-shaped} convolutional pattern to maximize receptive field coverage with minimal memory growth. After training, these components are converted into efficient LUT lookups, enabling color denoising with only $\sim$500\,KB of storage. Despite its tiny footprint, DnLUT surpasses prior LUT-based models by over 1~dB PSNR in denoising quality, while running $\sim$20× faster and consuming just 0.1\% of the energy of a standard CNN. This establishes a new state-of-the-art for resource-efficient image denoising using LUTs.

Researchers have also expanded LUT applications beyond super-resolution and denoising. \textit{Yang et al.} developed ICELUT~\cite{ICELUT}, which is the first purely LUT-based solution for real-time image enhancement. By converting a lightweight pointwise CNN and a split fully-connected layer into multi-dimensional LUTs, ICELUT achieves near state-of-the-art retouching results with extremely low latency. Notably, it runs an entire high-definition enhancement in $\sim$0.4\,ms on GPU (7\,ms on CPU), over an order of magnitude faster than conventional CNN models. In the video domain, \textit{He et al.} introduced a Multi-Frame Deformable LUT approach for compressed video quality enhancement~\cite{He2025MFDLUT}. Their method uses a small CNN to align and fuse multi-frame features, then converts these modules into LUT form during inference, achieving an excellent trade-off between restoration performance and runtime efficiency. Such advancements underscore the growing versatility of LUT-based operators, which continue to set new benchmarks in efficiency across various image and video restoration tasks.

\section{Anisotropic Pooling}
\label{sec:method}

This section develops an \emph{anisotropic pooling} strategy for LUT-realizable CNN-based image restoration.
Our motivation stems from the observation that the commonly used \emph{isotropic} fusion, simple averaging of predictions from rotated versions of a patch, fails to capture the inherently \emph{directional} nature of natural images.
Edges, line structures, and repeated textures exhibit preferred orientations; hence, an effective pooling rule should adapt its fusion weights according to local orientation cues rather than treating all orientations as equally informative.
We first formalize the multi-orientation processing pipeline and highlight why isotropic averaging under-utilizes orientation diversity (see Fig.~\ref{fig:average} and Fig.~\ref{fig:pixel}).
Subsequently, we propose a robust, differentiable \emph{generalized median pooling} (GMP) method (see Fig.~\ref{fig:gmp}) to mitigate outlier effects, followed by a fully \emph{learned orientation-aware pooling} (OAP) module that predicts content-adaptive fusion weights via a compact LUT-realizable CNN and integrates seamlessly into standard LUT-based image restoration frameworks (see Fig.~\ref{fig:oap}).

\subsection{Multi-Orientation Forward Model and Notation}
\label{subsec:unified}

Let $\mathbf{p}\in\mathbb{R}^{n}$ denote a vectorized local patch (e.g., $n=9$ for $3\times3$ neighborhoods). 
Let $\{R_i\}_{i=1}^k$ be a fixed set of planar symmetries (rotations or flips) acting linearly on patches; we use $k=4$ (rotations by $0^\circ,90^\circ,180^\circ,270^\circ$) unless otherwise stated. 
Let $f:\mathbb{R}^{n}\!\to\!\mathbb{R}^{m}$ denote a \emph{LUT-realizable} restoration operator (e.g., a super-resolution LUT queried via multilinear interpolation). 
For each orientation $i$, we rotate the input patch, perform restoration using the LUT, and then invert the rotation:
\begin{equation}
\label{eq:xi}
\mathbf{x}_i 
\;=\; 
R_i^{-1}\!\big(f(R_i(\mathbf{p}))\big)
\;\in\; 
\mathbb{R}^{m},
\qquad i=1,\dots,k .
\end{equation}

The orientation-specific predictions $\{\mathbf{x}_i\}_{i=1}^{k}$ are then aggregated into a final output through a convex combination:
\begin{equation}
\label{eq:pooling_general}
\begin{aligned}
\hat{\mathbf{y}}(\mathbf{p}) 
&= \sum_{i=1}^{k} \alpha_i(\mathbf{p})\, \mathbf{x}_i,\\
\text{s.t.}\quad 
&\alpha_i(\mathbf{p}) \ge 0,\quad 
\sum_{i=1}^{k}\alpha_i(\mathbf{p}) = 1 .
\end{aligned}
\end{equation}

Here, the weights $\boldsymbol{\alpha}(\mathbf{p})$ determine how much each orientation contributes to the final estimate. 
Conventional \emph{average pooling} assumes uniform confidence across orientations, fixing $\alpha_i(\mathbf{p}) \equiv 1/k$. 
In contrast, our goal is to learn adaptive weights $\boldsymbol{\alpha}(\mathbf{p})$ that reflect the reliability of each orientation—capturing, for example, texture anisotropy or structural consistency inferred directly from the local patch $\mathbf{p}$.

\subsection{Drawbacks of Isotropic Averaging}
\label{subsec:avg_drawbacks}

\begin{figure}[t]
    \centering
    \includegraphics[width=0.99\linewidth]{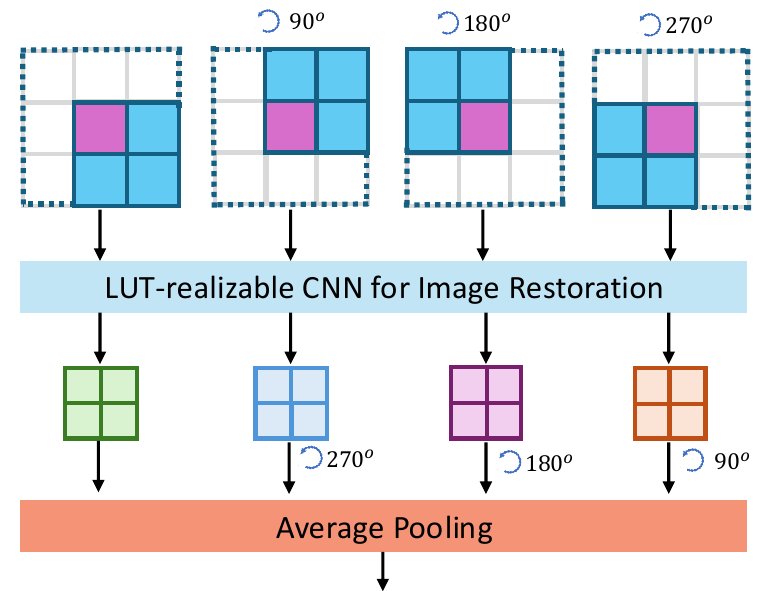}
    \caption{
    Illustration of the average pooling mechanism in a LUT-realizable CNN for image restoration. 
    A \(2\times2\) input patch is rotated by \(0^\circ\), \(90^\circ\), \(180^\circ\), and \(270^\circ\), processed by the network, 
    and then inversely rotated before being merged through average pooling. 
    This common averaging operation effectively expands the receptive field to \(3\times3\), 
    but implicitly assumes that all orientations contribute equally.
    }
    \label{fig:average}
\end{figure}

Figure~\ref{fig:average} summarizes the conventional practice: a $2\times2$ input patch is rotated by $0^\circ/90^\circ/180^\circ/270^\circ$, each rotated version is processed by the LUT-realizable network, the outputs are inverse-rotated, and the four results are \emph{averaged}. 
This trick effectively expands the receptive field from $2\times2$ to $3\times3$ at negligible cost and is therefore pervasive in LUT-based restoration pipelines. 
However, this construction silently enforces an \emph{isotropy} assumption: all orientations are treated as equally informative. 
Natural images rarely satisfy this assumption. 
When a strong edge runs at $\sim 45^\circ$, the orientation whose receptive field aligns with that edge typically captures sharper high-frequency content than the one orthogonal to it; when a patch straddles a boundary, some orientations look across heterogeneous regions while others remain within a coherent area. 
Averaging \emph{dilutes} the most informative orientation with less relevant ones and is fragile to outliers.

\begin{figure}[t]
    \centering
    \includegraphics[width=0.99\linewidth]{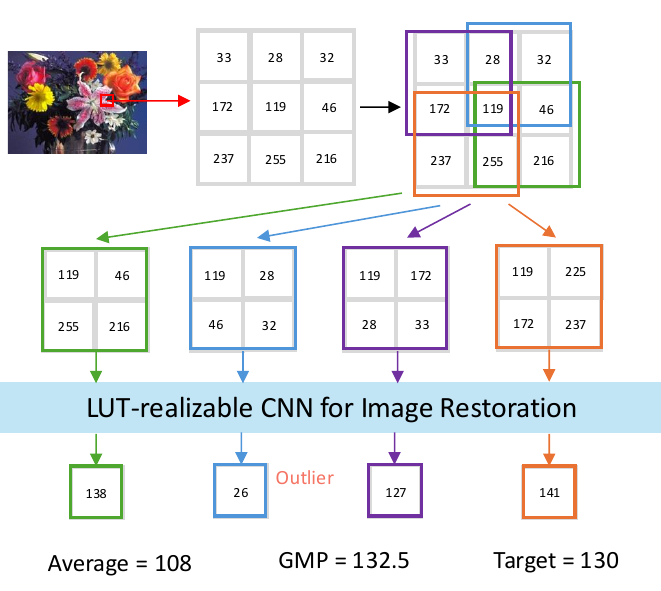}
    \caption{
    Illustration of anisotropic signal structures within a $3\times3$ image patch (extracted from the red square). 
    Different orientations produce distinct estimates, showing that simple averaging (108) can yield suboptimal results when outliers (56) distort the prediction. 
    In contrast, the proposed generalized median pooling (GMP = 132.5) provides a more robust and accurate estimate, closely matching the target value (= 130).
    }
    \label{fig:pixel}
\end{figure}

The numerical toy in Fig.~\ref{fig:pixel} makes this concrete. 
There, the four oriented predictions lead to different scalar estimates due to anisotropic structure within the $3\times3$ neighborhood extracted from the red box. 
The simple average value (108) deviates markedly from the target value (130) because one orientation (56) acts as an outlier. 
This illustrates two desiderata for a better pooling rule, which our methods pursue explicitly: 
(i) \emph{robustness} to noisy orientations, and 
(ii) \emph{adaptivity} to local structures so that orientations aligned with local structures can achieve higher weights.

\subsection{Generalized Median Pooling (GMP)}
\label{subsec:gmp}

The fundamental limitation of average pooling lies in its \emph{sensitivity to outliers}.  
When multiple orientation-specific predictions $\{\mathbf{x}_i\}_{i=1}^{k}$ are averaged with equal weights, even one inaccurate orientation can distort the final output, especially when that orientation corresponds to a patch crossing a strong edge or texture discontinuity.  
For instance, if three orientations give consistent high-intensity predictions while one gives a much lower value (e.g., due to mixing pixels from two regions), the mean will be pulled downward, leading to a blurred or attenuated edge.  
This issue is particularly critical in LUT-based pipelines, where the number of orientations is small ($k=4$), so each orientation’s influence is significant.  

To counter this, we introduce a \emph{generalized median pooling (GMP)} strategy designed to be both \emph{robust} and \emph{differentiable}.  
In the idealized case of scalar values, the most robust aggregation rule is the median, which completely ignores extreme outliers.  
However, the classical “hard median’’ involves sorting and selecting middle elements, which is inherently non-differentiable and thus unsuitable for gradient-based optimization.  
We therefore design a \emph{soft-median} that behaves similarly to the median—emphasizing values near the consensus while softly suppressing distant ones—but can be seamlessly integrated into end-to-end training.

Let the (channel-wise) mean of oriented predictions be
\begin{equation}
\label{eq:xbar}
\bar{\mathbf{x}} \;=\; \frac{1}{k}\sum_{i=1}^{k} \mathbf{x}_i .
\end{equation}
We first measure how far each orientation deviates from the consensus by computing the distance
\begin{equation}
d_i = \|\mathbf{x}_i - \bar{\mathbf{x}}\|,
\end{equation}
where the $L_1$ or $L_2$ norm is used over output channels.  
Intuitively, if an orientation prediction $\mathbf{x}_i$ is very different from the average, it is likely an \emph{outlier} and should contribute less to the final output.  
To realize this behavior smoothly, we transform the distances into normalized soft weights using a softmin function with a temperature parameter $\tau>0$:
\begin{equation}
\label{eq:gmp_weights}
\alpha_i^{\text{GMP}} 
= 
\frac{\exp(-d_i/\tau)}{\sum_{j=1}^{k}\exp(-d_j/\tau)}
\;\;\;\Rightarrow\;\;\;
\mathbf{y}^{\text{GMP}}
=
\sum_{i=1}^{k}\alpha_i^{\text{GMP}}\,\mathbf{x}_i
\end{equation}
The temperature $\tau$ controls how aggressively we suppress outliers:
as $\tau\!\to\!\infty$, all weights become equal ($\alpha_i^{\text{GMP}}\!=\!1/k$), reducing to average pooling;  
as $\tau\!\to\!0^{+}$, the weights concentrate on the predictions closest to the mean, mimicking the behavior of a hard median that discards extreme values.

\begin{figure}[t]
    \centering
    \includegraphics[width=0.99\linewidth]{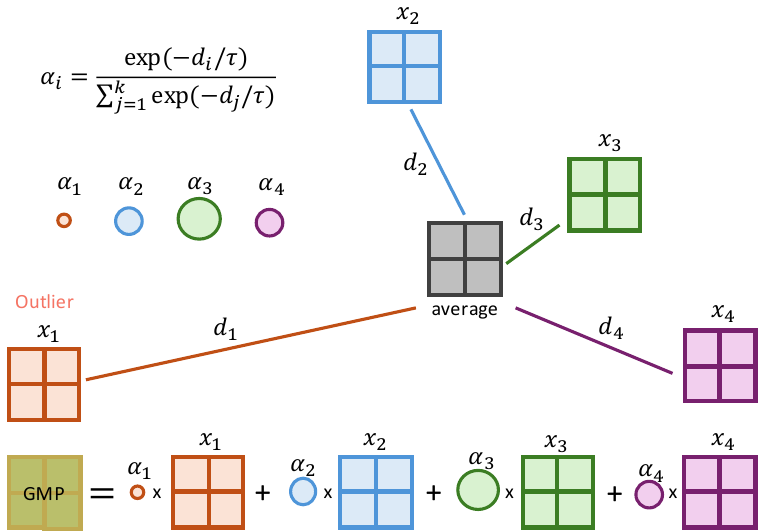}
    \caption{
    Illustration of the proposed generalized median pooling (GMP) strategy. 
    Patches whose values are closer to the average receive higher weights, while outliers are downweighted, 
    resulting in a robust, median-like aggregation that suppresses anomalies and preserves structural consistency.
    }
    \label{fig:gmp}
\end{figure}

Figure~\ref{fig:gmp} illustrates this mechanism.  
Predictions near the mean receive higher weights, while those far away are softly suppressed, resulting in a stable, median-like estimate that is less affected by erroneous orientations.  
This smooth weighting also makes the operator fully differentiable, allowing gradients to flow through both $\mathbf{x}_i$ and $\alpha_i^{\text{GMP}}$, enabling joint training of the LUT parameters and pooling temperature.

The practical advantage of GMP is that it effectively filters out “bad orientations’’ that would otherwise blur fine details or distort textures, while keeping the computational and storage cost almost identical to averaging.  
It provides a principled middle ground between rigid averaging and unstable hard selection.  
However, GMP still assumes that all orientations are \emph{a priori} equally important once their deviations are comparable, it does not explicitly learn which orientations should be emphasized depending on local geometry.  
For example, in an edge region aligned with $45^\circ$, the diagonal orientation consistently provides more reliable information than the others, but GMP treats all near-mean predictions equally.  
This motivates us to move one step further to a \emph{fully adaptive pooling} rule that can learn, from data, how much each orientation should contribute depending on the spatial structure.  
We term this approach \emph{Orientation-Aware Pooling (OAP)}, detailed in the next subsection.

\subsection{Learned Orientation-Aware Pooling (OAP)}
\label{subsec:oap}

GMP improves robustness by attenuating outliers, but it is still \emph{structure-agnostic}: once several orientations look “reasonable,” GMP treats them similarly and does not actively \emph{prefer} the one that best aligns with local geometry (e.g., the orientation parallel to an edge). 
To fully exploit directional cues, we make the fusion weights \emph{content-adaptive}—they should depend on the input patch itself rather than being fixed by a rule that is the same everywhere.

We therefore predict the weights in~\eqref{eq:pooling_general} directly from the input patch via a tiny coefficient LUT $C$:
\begin{equation}
\label{eq:oap_rule}
\boldsymbol{\alpha}(\mathbf{p}) \;=\; C(\mathbf{p}) \in \Delta^{k-1}
\;\;\Rightarrow\;\;
\hat{\mathbf{y}}(\mathbf{p}) \;=\; \sum_{i=1}^{k} \big[C(\mathbf{p})\big]_i \, \mathbf{x}_i
\end{equation}
where $\Delta^{k-1}$ is the probability simplex (nonnegative entries summing to one), $\mathbf{p}\!\in\!\mathbb{R}^{n}$ is the vectorized local patch, and $\mathbf{x}_i$ are the $k$ oriented predictions from~\eqref{eq:xi}. 
The mapping $C:\mathbb{R}^{n}\!\to\!\Delta^{k-1}$ is implemented as a small SR-LUT (queried via multilinear interpolation), so the entire pipeline remains LUT-realizable and integer-friendly at deployment. 
Intuitively, $C(\mathbf{p})$ learns a \emph{reliability profile} for orientations conditioned on the patch: orientations that look \emph{aligned} with local structures (edges, fine textures) receive higher weights; orientations that likely cut across boundaries or mix distinct regions are downweighted.

A complementary interpretation is to model each oriented estimate as $\mathbf{x}_i=\mathbf{y}+\boldsymbol{\epsilon}_i$, where $\boldsymbol{\epsilon}_i$ are orientation-dependent errors. 
If errors were independent with variances $\sigma_i^2$, the optimal linear fusion would weight inversely to $\sigma_i^2$. 
OAP learns a data-driven surrogate of these inverse-variance weights \emph{without} explicitly estimating variances: the coefficient LUT $C(\mathbf{p})$ infers which orientations are likely to be low-error from the observed patch statistics.

\begin{figure}[t]
    \centering
    \includegraphics[width=0.99\linewidth]{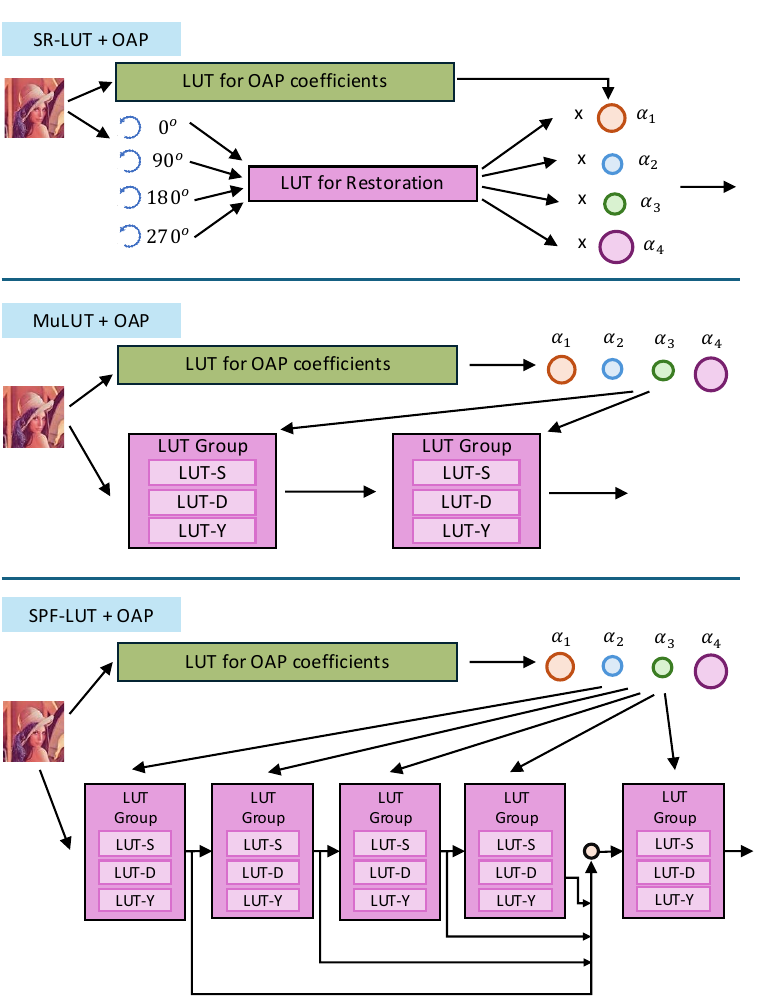}
    \caption{
    Overview of the proposed OAP (referenced in Sec.~\ref{subsec:oap}). 
    A tiny coefficient LUT $C(\mathbf{p})$ predicts orientation weights $\alpha_i$ from the input patch, which then fuse the inverse-rotated predictions from $k$ orientations. 
    The mechanism favors orientations aligned with local structures while preserving LUT efficiency.
    }
    \label{fig:oap}
\end{figure}

Figure~\ref{fig:oap} shows OAP integrated into three representative LUT pipelines:
(i) \textit{SR-LUT}: OAP follows a single restoration LUT and fuses the $k$ inverse-rotated outputs with the learned $\boldsymbol{\alpha}(\mathbf{p})$;
(ii) \textit{MuLUT}: coordinated LUTs first produce oriented predictions using complementary indexing; OAP then reweights orientations to emphasize the most informative directions;
(iii) \textit{SPF-LUT} cascade: we query $C(\mathbf{p})$ \emph{once} and \emph{reuse} the same $\boldsymbol{\alpha}(\mathbf{p})$ across stages (default), or re-estimate per stage when extra adaptivity is desired.
In all cases, $C(\cdot)$ can be evaluated \emph{in parallel} with the $k$ restoration queries, so the measured wall-clock overhead is negligible (see Sec.~\ref{sec:experiments}).

Both the restoration LUT $f(\cdot)$ and the coefficient LUT $C(\cdot)$ are queried by multilinear interpolation. 
Each query is a smooth weighted sum of nearby table entries; consequently, the mapping $\mathbf{p}\mapsto\hat{\mathbf{y}}(\mathbf{p})$ in~\eqref{eq:oap_rule} is piecewise linear and differentiable almost everywhere. 
Given training data $\mathcal{D}=\{(\mathbf{p}_t,\mathbf{y}_t)\}$, we optimize the LUT entries of $f$ and $C$ jointly by minimizing
\begin{equation}
\label{eq:loss}
\mathcal{L}(\Theta) 
\;=\;
\sum_{t} \ell\!\left(\hat{\mathbf{y}}(\mathbf{p}_t;\Theta), \mathbf{y}_t\right)
\;+\;
\lambda\,\mathcal{R}\!\big(\boldsymbol{\alpha}(\mathbf{p}_t)\big),
\end{equation}
where $\Theta$ collects all LUT parameters, $\ell$ is a fidelity loss (Charbonnier/$L_1$/$L_2$), and $\mathcal{R}$ regularizes the weight distribution (e.g., entropy to avoid collapse, or temperature-like priors to control sharpness). 
Gradients propagate (i) to LUT entries via interpolation weights, and (ii) to $\boldsymbol{\alpha}(\mathbf{p})$ through the simplex parameterization inside $C$, making the whole system trainable end-to-end.

Let $T_f$ be the cost of one restoration LUT query and $T_C$ the cost of the coefficient LUT. 
Average pooling costs $kT_f$; GMP adds a light distance computation; OAP adds $T_C$, yielding $kT_f+T_C$. 
Because $T_C\!\ll\!T_f$ and $C(\cdot)$ runs in parallel with the $k$ calls to $f(\cdot)$, the net runtime overhead is negligible in practice. 
For storage, a LUT with sampling interval $q$, receptive-field dimension $n$, outputs per entry $m$, and bit depth $B$ requires
\begin{equation}
\label{eq:lut_size}
S \;=\; \bigl(2^{8-q}+1\bigr)^{n}\times m B .
\end{equation}
Our coefficient LUT uses a tiny configuration (e.g., $q{=}5$, $n{=}4$, $m{=}k$, $B{=}8$ bits), resulting in a pre-compression footprint on the order of tens of kilobytes and, with LUT compression (e.g., DFC), often only a few kilobytes—orders of magnitude smaller than the main restoration LUTs (see Sec.~\ref{sec:experiments}).

The set $\{R_i\}$ forms a discrete rotation/flip group. 
Average pooling enforces strict invariance to this group (all orientations equal), which can harm directional detail in anisotropic regions. 
GMP relaxes this by suppressing inconsistent orientations. 
OAP goes further and learns \emph{conditional invariance}: in flat or isotropic regions, $C(\mathbf{p})$ produces near-uniform weights; near structured edges or textures, it produces peaked weights that privilege the best-aligned orientations. 
This conditional behavior explains the consistent PSNR/SSIM gains we observe across tasks: OAP preserves sharp, orientation-sensitive content without sacrificing the efficiency that makes LUT-based methods attractive in the first place.

\subsection{Residual Learning}
\label{subsec:residual}

Residual learning simplifies the prediction target and stabilizes training. 
Let $\mathbf{x}_{\text{base}}$ be a cheap baseline (e.g., bicubic upsampling for SR). 
Instead of predicting the full output, the LUT-realizable module predicts a residual $\mathbf{r}$, and we form
\begin{equation}
\label{eq:residual}
\mathbf{y}_{\text{final}} \;=\; \mathbf{x}_{\text{base}} + \mathbf{r}.
\end{equation}
This reduces dynamic range, focuses capacity on high-frequency refinements, and empirically accelerates convergence while mitigating over-smoothing. 
In our experiments, residual learning yields consistent (albeit modest) PSNR/SSIM gains on top of both GMP and OAP, especially within multi-stage LUT cascades where stability matters.

\section{Experiments and Results}
\label{sec:experiments}

To comprehensively evaluate the effectiveness of our proposed \emph{anisotropic pooling} mechanisms, we conduct experiments on four representative image restoration tasks: super-resolution, denoising, deblocking, and deblurring. 
Our goal is to verify that the proposed generalized median pooling (GMP) and orientation-aware pooling (OAP) can be seamlessly integrated into existing LUT-based restoration frameworks, yielding consistent accuracy gains with minimal overhead.

\textbf{Experimental setup.} 
We build upon three representative LUT-based models, SR-LUT~\cite{SRLUT}, MuLUT~\cite{MULUT}, SPF-LUT~\cite{DFCLUT}, and train them using the widely adopted DIV2K~\cite{agustsson2017ntire} dataset for super-resolution, denoising, and deblocking tasks, and the GoPro dataset~\cite{nah2017deep} for motion deblurring. 
Following the standard adaptation procedure, we remove the PixelShuffle layer~\cite{li2024toward} for non-super-resolution tasks to ensure compatibility with the LUT-based inference structure. 
All models are trained for $2\times10^5$ iterations on a single NVIDIA V100 GPU (32GB) using the Adam optimizer with a cosine-annealed learning rate schedule, starting from $1\times10^{-4}$. 
We employ a batch size of 32 and use random $48\times48$ patch cropping with random rotations and flips for data augmentation to enhance generalization.

\textbf{Integration of anisotropic pooling.} 
To evaluate our proposed methods, we integrate GMP and OAP into the three LUT-based baselines, denoted as ``+GMP'' and ``+OAP'' variants, respectively. 
This integration replaces the conventional isotropic averaging step in multi-orientation fusion with our anisotropic pooling modules. 
After the main training stage, we perform a brief fine-tuning step following the strategy in~\cite{MULUT}, allowing the newly introduced pooling mechanism to co-adapt with the pre-trained LUTs in an end-to-end manner. 
This ensures that the orientation-sensitive pooling coefficients remain consistent with the learned feature statistics of the underlying LUTs.

\textbf{Evaluation protocol.}
We report results on standard benchmark datasets, including Set5, Set14, BSDS100, Urban100, and Manga109 for super-resolution, as well as Set12 and BSD68 for denoising.
For each task, we measure both PSNR and SSIM to assess reconstruction quality, and we also compare memory storage and computational overhead to evaluate efficiency. 
We emphasize that our method does not introduce any significant increase in model size or runtime, as the OAP module is implemented as a compact LUT that can be queried in parallel with the main restoration LUTs.

\subsection{Image super-resolution}
\begin{table*}[t]
\centering
\caption{
Quantitative comparison of PSNR/SSIM and model storage for $\times4$ image super-resolution. 
The \colorbox{blue!15}{blue} and \colorbox{orange!15}{orange} rows indicate LUT-based baselines 
enhanced with the proposed anisotropic pooling strategies: 
generalized median pooling (``\colorbox{blue!15}{+GMP}'') and orientation-aware pooling (``\colorbox{orange!15}{+OAP}''). 
Across all benchmark datasets, OAP consistently achieves the highest PSNR/SSIM gains with only a negligible increase in storage, demonstrating its strong balance between accuracy and efficiency.
}
\label{tab:sr}
\begin{tabular}{llcccccc}
\toprule
 & \textbf{Method} & \textbf{Storage} & \textbf{Set5} & \textbf{Set14} & \textbf{BSDS100} & \textbf{Urban100} & \textbf{Manga109} \\
\midrule
\multirow{4}{*}{\textbf{Classical}} 
 & Bicubic           
                & --       & 28.42/0.8101      & 26.00/0.7023      & 25.96/0.6672      & 23.14/0.6574      & 24.91/0.7871 \\
 & NE + LLE~\cite{chang2004super}  
                & 1.434MB  & 29.62/0.8404      & 26.82/0.7346      & 26.49/0.6970      & 23.84/0.6942      & 26.10/0.8195 \\
 & ANR~\cite{timofte2013anchored}       
                & 1.434MB  & 29.70/0.8422      & 26.86/0.7368      & 26.52/0.6992      & 23.89/0.6964      & 26.18/0.8214 \\
 & A+~\cite{timofte2015a+}        
                & 15.17MB  & 30.27/0.8602      & 27.30/0.7498      & 26.73/0.7088      & 24.33/0.7189      & 26.91/0.8480 \\
\midrule
\multirow{2}{*}{\textbf{DNN}} 
 & RRDB~\cite{wang2018esrgan} 
                & 63.942MB & 32.68/0.8999      & 28.88/0.7891      & 27.82/0.7444      & 27.02/0.8146     & 31.57/0.9185 \\
 & EDSR~\cite{lim2017enhanced} 
                & 164.396MB& 32.46/0.8968      & 28.80/0.7876      & 27.71/0.7420      & 26.64/0.8033     & 31.02/0.9148 \\
\midrule
\multirow{10}{*}{\textbf{LUT}}
 & SR-LUT~\cite{SRLUT}   
                & 1.274MB  & 29.94/0.8524      & 27.18/0.7416      & 26.59/0.6999      & 24.09/0.7053      & 26.94/0.8454 \\
 & SR-LUT~\cite{SRLUT} + DFC~\cite{DFCLUT} 
                & 0.128MB  & 29.88/0.8501      & 27.14/0.7394      & 26.57/0.6982      & 24.05/0.7021      & 26.87/0.8423 \\
\rowcolor{blue!15}
\cellcolor{white} & SR-LUT~\cite{SRLUT} + DFC~\cite{DFCLUT} + GMP 
                & 0.128MB  & 29.99/0.8542      & 27.25/0.7428      & 26.66/0.7002      & 24.15/0.7069      & 26.97/0.8481 \\
\rowcolor{orange!15}
\cellcolor{white} & SR-LUT~\cite{SRLUT} + DFC~\cite{DFCLUT} + OAP 
                & 0.131MB  & 30.18/0.8563      & 27.43/0.7465      & 26.87/0.7042      & 24.37/0.7095      & 27.17/0.8519 \\
\cmidrule(lr){2-8}
 & MuLUT~\cite{MULUT}
                & 4.062MB  & 30.60/0.8653      & 27.60/0.7541      & 26.86/0.7110      & 24.46/0.7194      & 27.90/0.8633 \\
 & MuLUT~\cite{MULUT} + DFC  
                & 0.407MB  & 30.55/0.8642      & 27.56/0.7532      & 26.83/0.7104      & 24.41/0.7177      & 27.82/0.8613 \\
\rowcolor{blue!15}
\cellcolor{white} & MuLUT~\cite{MULUT} + DFC + GMP 
                & 0.407MB  & 30.63/0.8668      & 27.68/0.7556      & 26.90/0.7119      & 24.52/0.7208      & 27.94/0.8647 \\
\rowcolor{orange!15}
\cellcolor{white} & MuLUT~\cite{MULUT} + DFC + OAP 
                & 0.410MB  & 30.84/0.8704      & 27.89/0.7591      & 27.11/0.7157      & 24.75/0.7236      & 28.11/0.8675 \\
\cmidrule(lr){2-8}
 & SPF-LUT~\cite{DFCLUT}        
                & 17.284MB & 31.11/0.8764      & 27.92/0.7640      & 27.10/0.7197      & 24.87/0.7378      & 28.68/0.8796 \\
 & SPF-LUT~\cite{DFCLUT} + DFC   
                & 2.018MB  & 31.05/0.8755      & 27.88/0.7632      & 27.08/0.7190      & 24.81/0.7357      & 28.58/0.8779 \\
\rowcolor{blue!15}
\cellcolor{white} & SPF-LUT~\cite{DFCLUT} + DFC + GMP 
                & 2.018MB  & 31.18/0.8778      & 27.99/0.7648      & 27.15/0.7209      & 24.90/0.7388      & 28.65/0.8807 \\
\rowcolor{orange!15}
\cellcolor{white} & SPF-LUT~\cite{DFCLUT} + DFC + OAP 
                & 2.021MB  & 31.39/0.8812      & 28.16/0.7681      & 27.36/0.7243      & 25.08/0.7412      & 28.84/0.8832 \\
\bottomrule
\end{tabular}
\end{table*}

\begin{figure*}[!h]
    \centering
    \includegraphics[width=0.92\linewidth]{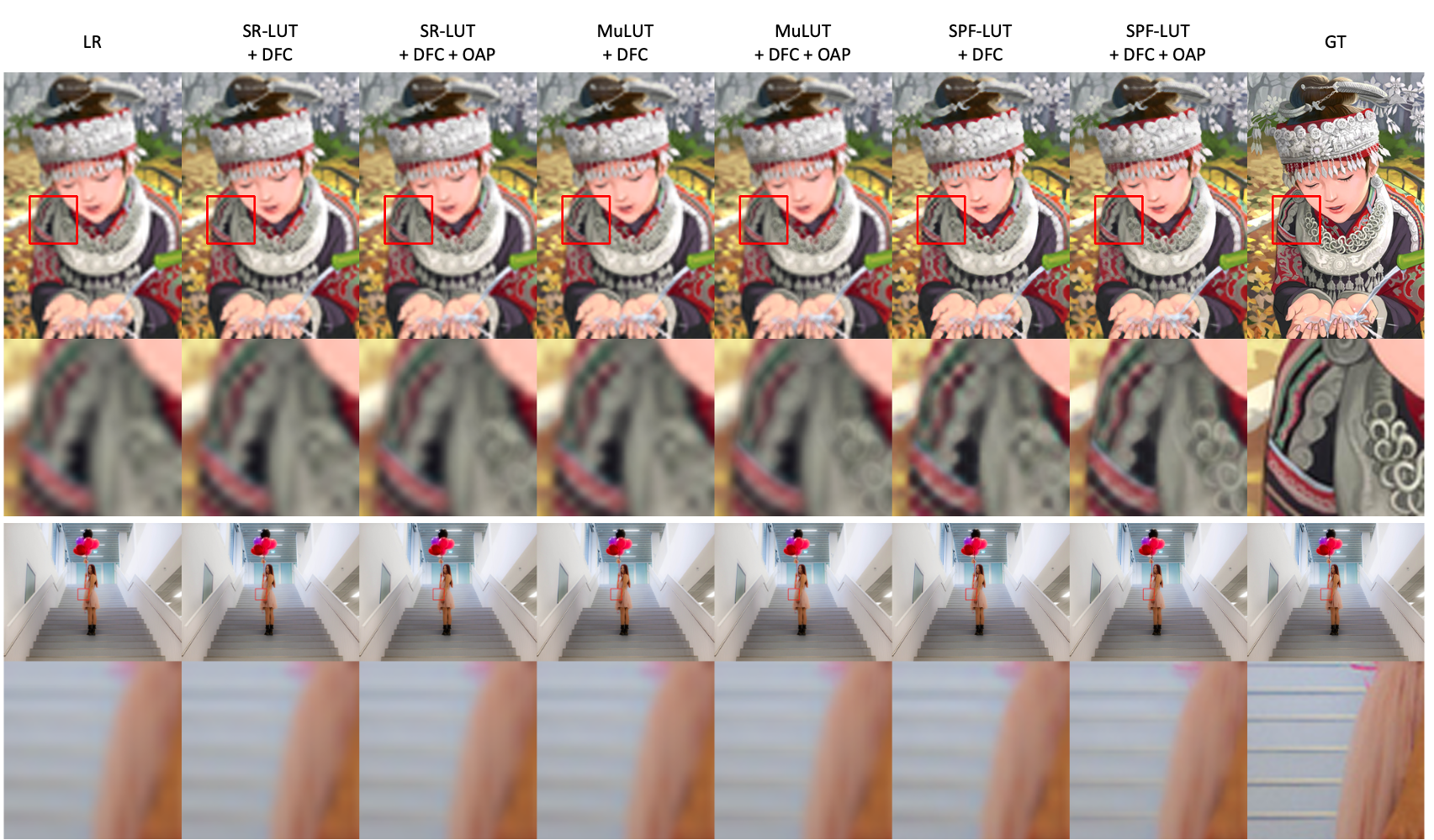}
    \caption{
        Visual comparisons of different LUT-based super-resolution methods for the $\times4$ task. 
        Compared to their baseline counterparts, the proposed OAP-equipped models
        recover sharper textures and clearer structural details, such as the clothing patterns and stair edges—while maintaining the lightweight efficiency characteristic of LUT-based frameworks.
        }
    \label{fig:visual}
\end{figure*}

We first evaluate our anisotropic pooling strategies (GMP and OAP) on the widely used $\times4$ image super-resolution benchmarks: Set5, Set14, BSDS100~\cite{martin2001database}, Urban100~\cite{huang2015single}, and Manga109~\cite{matsui2017sketch}. 
Low-resolution inputs are generated using bicubic downsampling, and performance is measured in terms of PSNR and SSIM on the luminance (Y) channel. 
For reference, we report results of both classical non-deep methods (Bicubic, NE+LLE~\cite{chang2004super}, ANR~\cite{timofte2013anchored}, A+~\cite{timofte2015a+}) and modern DNN-based models (RRDB~\cite{wang2018esrgan}, EDSR~\cite{lim2017enhanced}). 
Within the LUT family, we compare three representative baselines—SR-LUT~\cite{SRLUT}, MuLUT~\cite{MULUT}, and SPF-LUT~\cite{DFCLUT}—along with their DFC-compressed counterparts, against our proposed ``+GMP'' and ``+OAP'' variants that replace the naive averaging fusion with anisotropic pooling.

\textbf{Quantitative comparison.}
As summarized in Table~\ref{tab:sr}, both GMP and OAP yield consistent and notable improvements across all benchmark datasets.  
Incorporating our anisotropic pooling strategies into existing LUT-based pipelines raises PSNR/SSIM scores by up to $0.3$\,dB with negligible storage overhead (typically under 2\,KB increase).  
For example, SR-LUT + DFC achieves 29.88\,dB on Set5, while SR-LUT + DFC + OAP improves to 30.18\,dB; similarly, MuLUT + DFC + OAP increases from 30.55\,dB to 30.84\,dB on Set5, and SPF-LUT + DFC + OAP reaches 31.39\,dB, outperforming its baseline by 0.34\,dB on Set14.  
These gains are consistent across datasets with different texture distributions—fine-grained patterns in Manga109, repetitive structures in Urban100, and natural scenes in BSDS100—demonstrating the broad generalization ability of anisotropic pooling.

It is also worth noting that the proposed OAP consistently surpasses GMP across all benchmarks.  
While GMP improves robustness by suppressing orientation outliers, OAP further adapts orientation weights based on local content, leading to sharper texture reconstruction and fewer aliasing artifacts.  
Despite these improvements, OAP adds minimal computational cost and keeps model storage within the sub-megabyte range, preserving the hallmark efficiency of LUT-based inference.

\textbf{Qualitative analysis.}
Visual comparisons in Fig.~\ref{fig:visual} further validate the advantage of anisotropic pooling.  
Across various scenes, the OAP-enhanced models produce more faithful structural details and sharper edges, recovering textures such as clothing patterns, window frames, and stair edges that are partially blurred in the baseline outputs.  
These qualitative improvements align with the numerical gains, confirming that the learned orientation-aware fusion effectively preserves high-frequency components without introducing ringing or oversharpening.  
Importantly, this enhancement is achieved without any increase in model complexity or inference latency.

Overall, our empirical  results indicate that anisotropic pooling substantially enhances the representational power of LUT-realizable CNN super-resolution methods.  
By selectively emphasizing orientation-consistent information, both GMP and OAP close much of the performance gap between lightweight LUT models and more computationally intensive CNN-based methods, while maintaining extreme efficiency suitable for real-time and edge deployment.

\subsection{Image Denoising}
\begin{table}[t]
\centering
\caption{
Comparison of PSNR (dB) and storage size (KB) on standard benchmark datasets for grayscale image denoising at noise level 15.
The \colorbox{blue!15}{blue} and \colorbox{orange!15}{orange} backgrounds indicate LUT-based baselines that combined with the proposed anisotropic pooling strategie (``\colorbox{blue!15}{+GMP}'' and ``\colorbox{orange!15}{+OAP}''). 
}
\label{tab:denoising}
\begin{tabular}{lcccc}
\toprule
\textbf{Method} & \textbf{Storage Size} & \textbf{Set12} & \textbf{BSD68} \\
\midrule
BM3D       & --                 & 32.37 & 31.07 \\
WNNM         & --                 & 32.70 & 31.37 \\
TNRD       & --                 & 32.50 & 31.42 \\
\midrule
DnCNN       & 2.187MB         & 32.86 & 31.73 \\
FFDNet       & 1.932MB         & 32.75 & 31.69 \\
SwinIR      & 116.422MB          & 33.36 & 31.97 \\
\midrule
SR-LUT                               & 81.563KB & 30.42 & 29.78 \\
SR-LUT + DFC           & 8.172KB  & 30.39 & 29.76 \\
\rowcolor{blue!15}
SR-LUT + DFC + GMP      & 8.172KB & 30.48 & 29.87\\
\rowcolor{orange!15}
SR-LUT + DFC + OAP      & 10.732KB & 30.74 & 30.11\\
\midrule
MuLUT                                 & 489.381KB & 31.50 & 30.63 \\
MuLUT + DFC           & 49.031KB & 31.38 & 30.54 \\
\rowcolor{blue!15}
MuLUT + DFC + GMP       & 49.031KB & 31.52 & 30.69 \\
\rowcolor{orange!15}
MuLUT + DFC + OAP       & 51.591KB & 31.72 & 30.89\\
\midrule                            
SPF-LUT                               & 3017.849KB & 32.11 & 31.17 \\
SPF-LUT + DFC           & 595.926KB & 32.01 & 31.09 \\
\rowcolor{blue!15}
SPF-LUT + DFC + GMP    & 595.926KB & 32.15 & 31.22 \\
\rowcolor{orange!15}
SPF-LUT + DFC + OAP    & 598.486KB & 32.40 & 31.46\\
\bottomrule
\end{tabular}
\end{table}

We also evaluate the proposed anisotropic pooling strategies (GMP and OAP) on grayscale image denoising using Set12~\cite{zhang2017beyond} and BSD68~\cite{martin2001database} under a Gaussian noise level of 15. As shown in Table~\ref{tab:denoising}, both strategies consistently improve PSNR across all LUT-based baselines with negligible storage overhead. In particular, SR-LUT + DFC + OAP achieves 30.74/30.11\,dB on Set12/BSD68, outperforming the baseline by about 0.35\,dB, while SPF-LUT + DFC + OAP further boosts performance to 32.40/31.46\,dB, surpassing classical denoising algorithms (BM3D, WNNM, TNRD) and approaching the accuracy of lightweight CNN-based models (DnCNN, FFDNet). The observed improvements indicate that anisotropic pooling not only enhances robustness to random noise but also preserves local edge and texture information—OAP adaptively assigns higher weights to orientation-consistent responses, suppressing misaligned estimates and reducing over-smoothing effects. Compared with GMP, OAP achieves greater gains, showing that learning orientation-aware coefficients brings stronger generalization across varied noise patterns. Overall, the empirical results confirm that integrating anisotropic pooling significantly improves the denoising capability of LUT-based networks without significantly compromising their hallmark compactness and inference efficiency.

\subsection{Image Deblocking}
\begin{table}[t]
\centering
\caption{
Comparison of PSNR-B on benchmark datasets for image deblocking (QF=10). 
The \colorbox{blue!15}{blue} and \colorbox{orange!15}{orange} backgrounds indicate LUT-based baselines that combined with the proposed anisotropic pooling strategie (``\colorbox{blue!15}{+GMP}'' and ``\colorbox{orange!15}{+OAP}''). 
}
\label{tab:deblocking}
\begin{tabular}{lcccc}
\toprule
\textbf{Method} & \textbf{Storage Size} & \textbf{Classic5} & \textbf{LIVE1} \\
\midrule
JPEG   & --         & 25.21 & 25.33 \\
SA-DCT  & --    & 28.15 & 28.01 \\
\midrule
ARCNN & 415.812KB & 28.76 & 28.77 \\
SwinIR & 97.560MB  & 29.95 & 29.50 \\
\midrule

SR-LUT & 81.563KB & 27.58 & 27.69 \\
SR-LUT + DFC & 8.172KB & 27.55 & 27.64 \\
\rowcolor{blue!15}
SR-LUT + DFC + GMP & 8.172KB & 27.68 & 27.76\\
\rowcolor{orange!15}
SR-LUT + DFC + OAP & 10.732KB & 27. 87 & 27.93\\
\midrule

MuLUT & 489.381KB & 28.29 & 28.39 \\
MuLUT + DFC & 49.031KB & 28.24 & 28.33 \\
\rowcolor{blue!15}
MuLUT + DFC + GMP & 49.031KB & 28.36 & 28.49 \\
\rowcolor{orange!15}
MuLUT + DFC + OAP & 51.591KB & 28.59 & 28.71 \\
\midrule 

SPF-LUT & 3017.849KB & 28.63 & 28.62 \\
SPF-LUT + DFC & 595.926KB & 28.62 & 28.61 \\
\rowcolor{blue!15}
SPF-LUT + DFC + GMP & 595.926KB & 28.70 & 28.71 \\
\rowcolor{orange!15}
SPF-LUT + DFC + OAP & 598.486KB & 28.91 & 28.94 \\

\bottomrule
\end{tabular}
\end{table}
Table~\ref{tab:deblocking} presents PSNR-B comparisons on Classic5~\cite{foi2007pointwise} and LIVE1~\cite{sheikh2005live} under JPEG quality factor~10, where PSNR-B better reflects perceived blocking artifacts. Both anisotropic pooling variants consistently enhance all LUT-based baselines, with GMP yielding modest gains of about 0.1\,dB and OAP providing further 0.2--0.3\,dB improvement. For example, SPF-LUT + DFC + OAP achieves 28.91/28.94\,dB on Classic5/LIVE1, surpassing classical deblocking methods (JPEG, SA-DCT) and narrowing the gap to deep models such as ARCNN and SwinIR, despite its extremely compact footprint of less than 0.6\,MB. The improvements highlight that adaptive orientation weighting effectively suppresses block boundaries introduced by quantization, recovering smoother textures and more natural gradients without over-smoothing details. GMP primarily contributes by downweighting inconsistent directional responses, while OAP adaptively emphasizes edge-aligned orientations that correlate with underlying signal geometry. Importantly, these benefits come with negligible increases in LUT size (typically under 3\,KB), demonstrating that anisotropic pooling is an efficient and scalable enhancement for LUT-based deblocking pipelines, providing perceptual gains comparable to much larger DNN-based models.

\subsection{Image Deblurring}
\begin{table}[t]
\centering
\caption{
Comparison of PSNR/SSIM on the GoPro test set for image deblurring.
The \colorbox{blue!15}{blue} and \colorbox{orange!15}{orange} backgrounds indicate LUT-based baselines that combined with the proposed anisotropic pooling strategie (``\colorbox{blue!15}{+GMP}'' and ``\colorbox{orange!15}{+OAP}''). 
}
\label{tab:deblurring}
\begin{tabular}{lcc}
\toprule
\textbf{Method} & \textbf{Storage Size} & \textbf{GoPro (PSNR/SSIM)} \\
\midrule
Xu~\etal         & --          & 21.00/0.7410 \\
Kim \& Lee      & --          & 23.64/0.8239 \\
\midrule
Gong~\etal     & --          & 26.06/0.8632 \\
DBGAN             & 44.318MB    & 31.10/0.9420 \\
\midrule

SR-LUT            & 81.563KB    & 25.69/0.8598 \\
SR-LUT + DFC     & 8.172KB & 25.68/0.8592 \\
\rowcolor{blue!15}
SR-LUT + DFC + GMP & 8.172KB & 25.74/0.8610 \\
\rowcolor{orange!15}
SR-LUT + DFC + OAP & 10.732KB & 25.86/0.8621 \\
\midrule

MuLUT             & 489.381KB   & 25.74/0.8604 \\
MuLUT + DFC        & 49.031KB & 25.73/0.8604 \\
\rowcolor{blue!15}
MuLUT + DFC + GMP & 49.031KB & 25.80/0.8611\\
\rowcolor{orange!15}
MuLUT + DFC + OAP & 51.591KB & 25.91/0.8638\\
\midrule 

SPF-LUT                       & 3017.849KB  & 25.94/0.8640 \\
SPF-LUT +DFC                  & 595.926KB   & 25.92/0.8627 \\
\rowcolor{blue!15}
SPF-LUT + DFC + GMP & 595/926KB & 25.99/0.8652\\
\rowcolor{orange!15}
SPF-LUT + DFC + OAP & 598.486KB & 26.24/0.8671\\
\bottomrule
\end{tabular}
\end{table}
We further validate the proposed anisotropic pooling strategies (GMP and OAP) on the GoPro~\cite{nah2017deep} dataset for motion deblurring, comparing against classical algorithms (Xu~\etal~\cite{xu2013unnatural}, Kim and Lee~\cite{hyun2014segmentation}) and modern deep learning models (Gong~\etal~\cite{gong2017motion}, DBGAN~\cite{zhang2020deblurring}). As summarized in Table~\ref{tab:deblurring}, integrating GMP or OAP consistently improves both PSNR and SSIM across all LUT-based baselines, while maintaining an extremely compact model size. For instance, SPF-LUT + DFC + OAP achieves 26.24\,dB PSNR and 0.8671 SSIM, surpassing its DFC baseline by 0.32\,dB and outperforming all classical methods by a significant margin. Similarly, SR-LUT and MuLUT models gain about 0.15--0.25\,dB when equipped with OAP, demonstrating the general effectiveness of adaptive orientation fusion. These improvements highlight that anisotropic pooling enables the LUT-based frameworks to better handle spatially variant blur, dynamically emphasizing orientations aligned with motion direction and thus restoring sharper edges and textures. Despite its simplicity, OAP bridges much of the performance gap toward deep CNN-based deblurrers like DBGAN~\cite{zhang2020deblurring}, yet requires less than 1\,MB of total storage—over 40$\times$ smaller—thereby offering a compelling balance between accuracy, efficiency, and deployability for edge-oriented image restoration.

\subsection{Ablation Study on Sampling Interval}
\begin{table}[t]
\centering
\caption{
Ablation study on the impact of the sampling interval \(q\) for OAP-LUT. 
As \(q\) decreases from 5 to 3, the LUT complexity increases, yet the gains in PSNR/SSIM remain marginal, demonstrating that a higher sampling interval (i.e., a simpler LUT) is sufficient for effective OAP.
}

\label{tab:ablation}
\begin{tabular}{lccccc}
\toprule
\textbf{OAP-LUT} & \(\boldsymbol{q}\) & \textbf{Entries} & \textbf{Memory (MB)} & \textbf{PSNR} & \textbf{SSIM} \\
\midrule
Tiny   & 5 & \(9^4\)    & 0.025  & 30.18 & 0.8563 \\
Medium & 4 & \(17^4\)   & 0.318  & 30.19 & 0.8566 \\
Large  & 3 & \(33^4\)   & 4.525  & 30.20 & 0.8569 \\
\bottomrule
\end{tabular}
\end{table}
We further analyze the effect of the sampling interval \(q\) in the OAP coefficient-predicting LUT, which directly determines its granularity and memory footprint. As shown in Table~\ref{tab:ablation}, decreasing \(q\) (i.e., using finer sampling) significantly increases the LUT size, from only 0.025\,MB at \(q{=}5\) to 4.525\,MB at \(q{=}3\), yet yields almost no improvement in reconstruction quality (PSNR rises by only 0.02\,dB and SSIM by 0.0006). This observation suggests that orientation-specific pooling coefficients can be well approximated even under coarse quantization, and that the OAP module does not require high-resolution sampling to function effectively. In practice, a higher sampling interval (\(q{=}5\)) provides the best trade-off between accuracy and memory efficiency, keeping the coefficient LUT extremely lightweight without compromising performance. This reinforces the practicality of OAP as a low-cost, plug-and-play enhancement for LUT-realizable CNN image restoration frameworks.

\subsection{Ablation Study on Residual Learning}
\begin{table}[t]
\centering
\caption{
Ablation study on the effect of residual learning for $\times4$ image super-resolution. 
While residual learning yields only a slight quantitative improvement ($\sim$0.05\,dB in PSNR), it plays a key role in stabilizing the optimization process and accelerating convergence during training, particularly for deeper LUT cascades such as SPF-LUT with OAP integration.
}

\begin{tabular}{lcc}
\toprule
Method & Set5 & Set14 \\
\midrule
SPF-LUT+DFC & 31.05/0.8755 & 27.88/0.7632 \\
\midrule
SPF-LUT+DFC+GMP (w/o residual) &  31.13/0.8769 &  27.96/0.7644 \\
SPF-LUT+DFC+GMP (w/ residual) & 31.18/0.8778 &  27.99/0.7648  \\
\midrule
SPF-LUT+DFC+OAP (w/o residual) & 31.33/0.8805 & 28.13/0.7673 \\
SPF-LUT+DFC+OAP (w/ residual) & 31.39/0.8812 & 28.16/0.7681  \\
\bottomrule
\end{tabular}
\label{tab:residual_ablation}
\end{table}
We further examine the impact of residual learning on LUT-based super-resolution models. As shown in Table~\ref{tab:residual_ablation}, introducing residual connections provides consistent, albeit modest, quantitative improvements—around 0.05\,dB in PSNR and slight gains in SSIM for both GMP- and OAP-equipped variants. Beyond these small numerical increases, the main advantage lies in training stability: residual learning alleviates gradient vanishing and accelerates convergence, particularly in deeper LUT cascades such as SPF-LUT. By allowing each LUT to focus on refining local errors rather than reconstructing the entire signal, residual connections simplify optimization and improve the robustness of anisotropic pooling modules. Overall, residual learning serves as an effective auxiliary mechanism that enhances training dynamics and fine-detail preservation without adding any storage or inference cost.

\subsection{Running Time Analysis}
\begin{table}[t]
\centering
\caption{
Runtime comparison for grayscale image denoising at $256\times256$ and $512\times512$ resolutions.
}
\label{tab:runtime}
\begin{tabular}{llccc}
\toprule
& \multirow{2}{*}{\textbf{Method}} & \multirow{2}{*}{\textbf{Platform}} & \textbf{RunTime} & \textbf{RunTime} \\
& & & (256$\times$256) & (512$\times$512) \\
\midrule
\multirow{4}{*}{\textbf{LUT}}
& SR-LUT             & Mobile  & 7    & 21    \\
& SR-LUT + OAP              & Mobile  & 9    & 25    \\
\cmidrule(lr){2-5}
& MuLUT               & Mobile  & 26    & 99   \\
& MuLUT + OAP               & Mobile  & 27    & 102   \\
\midrule
\multirow{3}{*}{\textbf{Classical}}
& BM3D                & PC     & 2599  & 12481 \\
& WNNM                & PC     & 84734 & 352732\\
& TNRD                & PC     & 1140  & 1564  \\
\midrule
\multirow{3}{*}{\textbf{DNN}}
& DnCNN               & Mobile & 635    & 2497    \\
& FFDNet              & Mobile & 167    & 550   \\
& SwinIR              & Mobile & 94849  & 362082\\
\bottomrule
\end{tabular}
\end{table}

A key advantage of the proposed orientation-aware pooling (OAP) module is that it introduces virtually no additional computational overhead. Because the OAP coefficient prediction is performed in parallel with the main LUT-based restoration, the overall latency remains almost unchanged. Specifically, while the primary LUT retrieves the restored outputs for each rotated patch, the OAP LUT simultaneously generates the corresponding orientation weights, after which a lightweight fusion step produces the final output. As summarized in Table~\ref{tab:runtime}, the runtime differences between baseline and OAP-enhanced models are minimal—for instance, SR-LUT increases only from 7 ms to 9 ms and MuLUT from 26 ms to 27 ms at $256\times256$ resolution—whereas the performance gains (Sec.~\ref{sec:experiments}) are substantial. In contrast, classical methods like WNNM or BM3D are three to four orders of magnitude slower, and DNN models such as SwinIR demand hundreds to thousands of times more inference time. These results confirm that OAP preserves the hallmark efficiency of LUT-based pipelines while delivering adaptive, high-quality restoration.

\subsection{OAP Energy Consumption and Deployment}
\begin{table*}[ht]
\centering
\caption{We compare the efficiency and performance of different models by assessing their energy consumption and peak memory usage when generating 1280 $\times$ 720 high-quality images with $\times$4 super-resolution. Additionally, we provide an analysis of the storage requirements for the libraries that each model depends on.}
\label{tab:energy}
\setlength{\tabcolsep}{6pt}
\footnotesize

\begin{tabular}{llccccccccc}
\toprule
& \multirow{2}{*}{\textbf{Method}} & int8 & int8 & int32 & int32 & float32  & float32 & Energy  & Peak  
& Dependent Library  \\
& & Add. & Mul.& Add. & Mul. & Add. & Mul. & Cost (pJ) & Memory &  Size \\
\midrule
\multirow{1}{*}{\textbf{Classical}} & 
Bicubic & & & & & 14.7M & 14.7M & 67.8M & 2.3MB & numpy: 16.5MB \\
\midrule
\multirow{6}{*}{\textbf{LUT}} 
& SR-LUT & 15.8M & 0.1M & 28.6M & 22.3M & - & - & 72.5M & 47.0MB & numpy: 16.5MB  \\
& SR-LUT + DFC & 17.2M & 0.1M & 28.6M & 22.3M & - & - & 72.5M & 40.5MB & numpy: 16.5MB  \\
& SR-LUT + DFC + OAP & 19.4M & 0.1M & 28.6M & 22.3M & - & - & 75.9M & 41.4MB & numpy: 16.5MB  \\
\cmidrule(lr){2-11}
& MuLUT & 5.3M & 0.2M & 93.0M & 71.8M & - & - & 233.6M & 50.7MB & numpy: 16.5MB  \\
& MuLUT + DFC & 6.1M & 0.2M & 93.0M & 71.8M & - & - & 233.9M & 41.3MB & numpy: 16.5MB  \\
& MuLUT + DFC + OAP& 7.5M & 0.2M & 93.0M & 71.8M & - & - & 238.6M & 46.2MB & numpy: 16.5MB  \\
\cmidrule(lr){2-11}
& SPF-LUT & 222.9M & 1.0M & 390.1M & 301.5M & - & - & 980.5M & 65.4MB & numpy: 16.5MB  \\
& SPF-LUT + DFC & 225.6M & 1.0M & 390.1M & 301.5M & - & - & 981.5M & 45.9MB & numpy: 16.5MB  \\
& SPF-LUT + DFC + OAP & 228.3M & 1.0M & 390.1M & 301.5M & - & - & 982.8M & 48.4MB & numpy: 16.5MB  \\
\midrule
\multirow{2}{*}{\textbf{DNN}} 
& RRDB & - & - & - & - & 1.0T & 1.0T & 4.7T & 843.6MB & torch(CPU): 186.3MB \\
& EDSR & - & - & - & - & 2.9T & 2.9T & 13.3T & 2.3GB & torch(CPU): 186.3MB  \\
\bottomrule
\end{tabular}
\end{table*}

The proposed Orientation-Aware Pooling (OAP) module enhances image restoration quality while preserving the low computational cost that defines LUT-based models. Unlike deep neural networks (DNNs), which rely on large-scale floating-point operations, OAP operates entirely with integer arithmetic. It introduces a small SR-LUT coefficient predictor that performs lightweight integer additions and multiplications, thereby maintaining both computational and energy efficiency. This design ensures that the additional operations introduced by OAP have minimal impact on power consumption, memory footprint, and deployment feasibility.

\paragraph{Energy Consumption Analysis}
Table~\ref{tab:energy} compares the computational energy costs of various models. Although integrating OAP slightly increases the number of integer operations, it does not introduce any floating-point computations, thereby retaining the intrinsic efficiency of LUT-based processing. The energy cost increase is minimal—typically below 3.5\% compared to the corresponding +DFC baselines. For example, SR-LUT+DFC consumes 72.5M pJ, while SR-LUT+DFC+OAP requires 75.9M pJ (+3.4\%). Similarly, MuLUT+DFC rises from 233.9M pJ to 238.6M pJ (+2.0\%), and SPF-LUT+DFC increases marginally from 981.5M pJ to 982.8M pJ (+0.1\%). In contrast, modern DNN-based models such as RRDB and EDSR require up to 13.3T pJ—several orders of magnitude higher due to their reliance on terascale floating-point multiplications. These comparisons highlight that OAP delivers improved adaptivity with negligible additional energy consumption.

\paragraph{Peak Memory and Deployment Considerations}
OAP also introduces only a minor memory overhead compared to LUT-based baselines. Because LUT models avoid the large intermediate feature maps required by DNNs, their peak memory usage remains low even when augmented with OAP. Specifically, SR-LUT+DFC increases from 40.5 MB to 41.4 MB (+2.2\%), MuLUT+DFC from 41.3 MB to 46.2 MB (+11.9\%), and SPF-LUT+DFC from 45.9 MB to 48.4 MB (+5.4\%). In contrast, DNN counterparts demand substantially more memory—RRDB requires 843.6 MB and EDSR 2.3 GB—representing 17$\times$ and 48$\times$ higher peaks, respectively. This sharp difference underscores the practicality of OAP-augmented LUT frameworks for deployment on devices with strict memory budgets.

\paragraph{Software and Deployment Feasibility}
From a software standpoint, LUT-based models with OAP are significantly easier to deploy than DNN-based networks. DNN frameworks such as PyTorch require extensive runtime libraries and dependencies, with CPU-only installations occupying around 186 MB, and additional GPU components further increasing the footprint. In contrast, LUT-based implementations can run efficiently with minimal dependencies—typically requiring only a lightweight NumPy-based environment of about 16 MB—or can even be implemented directly in C++ or Java for embedded systems. This minimal software stack enables deployment on mobile or IoT devices where both storage and compute resources are constrained.

\section{Conclusion}
\label{sec:conclusion}
In this paper, we introduced an anisotropic pooling framework to address the shortcomings of naive averaging in LUT-realizable CNN image restoration. We first highlighted how simple averaging can underexploit orientation-specific information, leading to blurring and diminished detail. To mitigate these effects, we proposed two key pooling strategies: generalized median pooling (GMP), which offers a more robust alternative to mean-based fusion, and learned orientation-aware pooling (OAP), which adaptively weighs the contributions of differently oriented pixel patches. Through extensive experiments on diverse restoration tasks, we demonstrated that both GMP and OAP consistently outperform existing LUT-based methods in terms of sharpness, quantitative scores, and overall efficiency. We believe the introduced anisotropic pooling strategies can serve as a powerful extension for future LUT-realizable CNN designs, paving the way for further innovations in efficient, high-quality image restoration.

\bibliographystyle{IEEEtran}
\bibliography{oap-lut}

\begin{thebibliography}{10}
\providecommand{\url}[1]{#1}
\csname url@samestyle\endcsname
\providecommand{\newblock}{\relax}
\providecommand{\bibinfo}[2]{#2}
\providecommand{\BIBentrySTDinterwordspacing}{\spaceskip=0pt\relax}
\providecommand{\BIBentryALTinterwordstretchfactor}{4}
\providecommand{\BIBentryALTinterwordspacing}{\spaceskip=\fontdimen2\font plus
\BIBentryALTinterwordstretchfactor\fontdimen3\font minus \fontdimen4\font\relax}
\providecommand{\BIBforeignlanguage}[2]{{%
\expandafter\ifx\csname l@#1\endcsname\relax
\typeout{** WARNING: IEEEtran.bst: No hyphenation pattern has been}%
\typeout{** loaded for the language `#1'. Using the pattern for}%
\typeout{** the default language instead.}%
\else
\language=\csname l@#1\endcsname
\fi
#2}}
\providecommand{\BIBdecl}{\relax}
\BIBdecl

\bibitem{SRLUT}
Y.~Jo and S.~J. Kim, ``Practical single-image super-resolution using look-up table,'' in \emph{Proceedings of the IEEE/CVF Conference on Computer Vision and Pattern Recognition}, 2021, pp. 691--700.

\bibitem{MULUT}
J.~Li, C.~Chen, Z.~Cheng, and Z.~Xiong, ``Mulut: Cooperating multiple look-up tables for efficient image super-resolution,'' in \emph{European conference on computer vision}.\hskip 1em plus 0.5em minus 0.4em\relax Springer, 2022, pp. 238--256.

\bibitem{SPLUT}
C.~Ma, J.~Zhang, J.~Zhou, and J.~Lu, ``Learning series-parallel lookup tables for efficient image super-resolution,'' in \emph{European Conference on Computer Vision}.\hskip 1em plus 0.5em minus 0.4em\relax Springer, 2022, pp. 305--321.

\bibitem{RCLUT}
G.~Liu, Y.~Ding, M.~Li, M.~Sun, X.~Wen, and B.~Wang, ``Reconstructed convolution module based look-up tables for efficient image super-resolution,'' in \emph{Proceedings of the IEEE/CVF International Conference on Computer Vision}, 2023, pp. 12\,217--12\,226.

\bibitem{DFCLUT}
Y.~Li, J.~Li, and Z.~Xiong, ``Look-up table compression for efficient image restoration,'' in \emph{Proceedings of the IEEE/CVF Conference on Computer Vision and Pattern Recognition}, 2024, pp. 26\,016--26\,025.

\bibitem{li2025tinylut}
H.~Li, J.~Guan, L.~Rui, S.~Ma, and L.~Gu, ``Tinylut: Tiny look-up table for efficient image restoration at the edge,'' \emph{Advances in Neural Information Processing Systems}, vol.~37, pp. 85\,340--85\,359, 2025.

\bibitem{dong2015compression}
C.~Dong, Y.~Deng, C.~C. Loy, and X.~Tang, ``Compression artifacts reduction by a deep convolutional network,'' in \emph{Proceedings of the IEEE international conference on computer vision}, 2015, pp. 576--584.

\bibitem{zhang2017beyond}
K.~Zhang, W.~Zuo, Y.~Chen, D.~Meng, and L.~Zhang, ``Beyond a gaussian denoiser: Residual learning of deep cnn for image denoising,'' \emph{IEEE transactions on image processing}, vol.~26, no.~7, pp. 3142--3155, 2017.

\bibitem{zhang2018ffdnet}
K.~Zhang, W.~Zuo, and L.~Zhang, ``Ffdnet: Toward a fast and flexible solution for cnn-based image denoising,'' \emph{IEEE Transactions on Image Processing}, vol.~27, no.~9, pp. 4608--4622, 2018.

\bibitem{wang2018esrgan}
X.~Wang, K.~Yu, S.~Wu, J.~Gu, Y.~Liu, C.~Dong, Y.~Qiao, and C.~Change~Loy, ``Esrgan: Enhanced super-resolution generative adversarial networks,'' in \emph{Proceedings of the European conference on computer vision (ECCV) workshops}, 2018, pp. 0--0.

\bibitem{zhang2020deblurring}
K.~Zhang, W.~Luo, Y.~Zhong, L.~Ma, B.~Stenger, W.~Liu, and H.~Li, ``Deblurring by realistic blurring,'' in \emph{Proceedings of the IEEE/CVF conference on computer vision and pattern recognition}, 2020, pp. 2737--2746.

\bibitem{agdl}
X.~Zhang and X.~Wu, ``Attention-guided image compression by deep reconstruction of compressive sensed saliency skeleton,'' in \emph{Proceedings of the IEEE/CVF Conference on Computer Vision and Pattern Recognition}, 2021, pp. 13\,354--13\,364.

\bibitem{davd}
X.~Zhang, X.~Wu, X.~Zhai, X.~Ben, and C.~Tu, ``Davd-net: Deep audio-aided video decompression of talking heads,'' in \emph{Proceedings of the IEEE/CVF Conference on Computer Vision and Pattern Recognition}, 2020, pp. 12\,335--12\,344.

\bibitem{mdvd}
X.~Zhang and X.~Wu, ``Multi-modality deep restoration of extremely compressed face videos,'' \emph{IEEE Transactions on Pattern Analysis and Machine Intelligence}, vol.~45, no.~2, pp. 2024--2037, 2022.

\bibitem{zhang2023lvqac}
------, ``Lvqac: Lattice vector quantization coupled with spatially adaptive companding for efficient learned image compression,'' in \emph{Proceedings of the IEEE/CVF Conference on Computer Vision and Pattern Recognition}, 2023, pp. 10\,239--10\,248.

\bibitem{zhang2024learning}
------, ``Learning optimal lattice vector quantizers for end-to-end neural image compression,'' in \emph{Advances in Neural Information Processing Systems}, vol.~37, 2024, pp. 106\,497--106\,518.

\bibitem{shi2025vmambair}
Y.~Shi, B.~Xia, X.~Jin, X.~Wang, T.~Zhao, X.~Xia, X.~Xiao, and W.~Yang, ``Vmambair: Visual state space model for image restoration,'' \emph{IEEE Transactions on Circuits and Systems for Video Technology}, 2025.

\bibitem{xu2025multirate}
H.~Xu, X.~Wu, and X.~Zhang, ``Multirate neural image compression with adaptive lattice vector quantization,'' in \emph{Proceedings of the Computer Vision and Pattern Recognition Conference}, 2025, pp. 7633--7642.

\bibitem{zeng2025low}
X.~Zeng, L.~Zhu, W.~Yang, H.~Leung, S.~Wang, and S.~Kwong, ``Low-light image enhancement via diffusion models with semantic priors of any region,'' \emph{IEEE Transactions on Circuits and Systems for Video Technology}, 2025.

\bibitem{zhang2025dc}
G.~Zhang, W.~Fang, Y.~Shang, Y.~Zheng, and W.~Lin, ``Dc 2 mnet: Lightweight and efficient discrete cosine channel modulation network for image restoration,'' \emph{IEEE Transactions on Circuits and Systems for Video Technology}, 2025.

\bibitem{vien2025gradient}
A.~G. Vien, J.~Lee, S.~Jeong, N.~Yang, C.~Lee \emph{et~al.}, ``Gradient-guided diffusion-based restoration of extremely compressed backgrounds for video coding for machines,'' \emph{IEEE Transactions on Circuits and Systems for Video Technology}, 2025.

\bibitem{Mantiuk2008}
R.~K. Mantiuk, S.~Daly, and L.~Kerofsky, ``Display adaptive tone mapping,'' \emph{ACM Transactions on Graphics}, vol.~27, no.~3, pp. 68:1--68:10, 2008.

\bibitem{de2008reducing}
D.~De~Caro, N.~Petra, and A.~G. Strollo, ``Reducing lookup-table size in direct digital frequency synthesizers using optimized multipartite table method,'' \emph{IEEE Transactions on Circuits and Systems I: Regular Papers}, vol.~55, no.~7, pp. 2116--2127, 2008.

\bibitem{Lefkimmiatis2009}
S.~Lefkimmiatis and M.~Unser, ``Real-time image denoising by spreading look-up tables,'' \emph{IEEE Transactions on Image Processing}, vol.~18, no.~11, pp. 2601--2614, 2009.

\bibitem{lee2010new}
J.-Y. Lee, J.-J. Lee, and S.~Park, ``New lookup tables and searching algorithms for fast h. 264/avc cavlc decoding,'' \emph{IEEE transactions on circuits and systems for video technology}, vol.~20, no.~7, pp. 1007--1017, 2010.

\bibitem{Pouli2011}
T.~Pouli and E.~Reinhard, ``Progressive color transfer for images of arbitrary dynamic range,'' \emph{Computers \& Graphics}, vol.~35, no.~1, pp. 67--80, 2011.

\bibitem{Rashid2011}
M.~H. Rashid, M.~K. Khan, and M.~I. Sarfraz, ``Efficient color enhancement technique for image and video using 3d look-up tables,'' \emph{International Journal of Computer Science and Network Security}, vol.~11, no.~2, pp. 93--98, 2011.

\bibitem{zeng2020learning}
H.~Zeng, J.~Cai, L.~Li, Z.~Cao, and L.~Zhang, ``Learning image-adaptive 3d lookup tables for high performance photo enhancement in real-time,'' \emph{IEEE Transactions on Pattern Analysis and Machine Intelligence}, vol.~44, no.~4, pp. 2058--2073, 2020.

\bibitem{zhang2022clut}
F.~Zhang, H.~Zeng, T.~Zhang, and L.~Zhang, ``Clut-net: Learning adaptively compressed representations of 3dluts for lightweight image enhancement,'' in \emph{Proceedings of the 30th ACM International Conference on Multimedia}, 2022, pp. 6493--6501.

\bibitem{li2024toward}
J.~Li, C.~Chen, Z.~Cheng, and Z.~Xiong, ``Toward dnn of luts: Learning efficient image restoration with multiple look-up tables,'' \emph{IEEE Transactions on Pattern Analysis and Machine Intelligence}, 2024.

\bibitem{AutoLUT}
Y.~Xu, S.~Yang, X.~Liu, J.~Liu, J.~Tang, and G.~Wu, ``{AutoLUT}: {LUT}-based image super-resolution with automatic sampling and adaptive residual learning,'' in \emph{Proceedings of the IEEE/CVF Conference on Computer Vision and Pattern Recognition (CVPR)}, 2025.

\bibitem{DnLUT}
S.~Yang, B.~Huang, Y.~Zhang, D.~Yu, Y.~Yang, and N.~Wong, ``{DnLUT}: Ultra-efficient color image denoising via channel-aware lookup tables,'' in \emph{Proceedings of the IEEE/CVF Conference on Computer Vision and Pattern Recognition (CVPR)}, 2025.

\bibitem{ICELUT}
S.~Yang, B.~Huang, M.~Cao, Y.~Ji, H.~Guo, N.~Wong, and Y.~Yang, ``Taming lookup tables for efficient image retouching,'' in \emph{Proceedings of the European Conference on Computer Vision (ECCV)}, 2024, pp. 144--159.

\bibitem{He2025MFDLUT}
G.~He, G.~Quan, C.~Wu, S.~Wang, D.~Zhou, and Y.~Li, ``Multi-frame deformable look-up table for compressed video quality enhancement,'' in \emph{Proceedings of the AAAI Conference on Artificial Intelligence}, 2025.

\bibitem{agustsson2017ntire}
E.~Agustsson and R.~Timofte, ``Ntire 2017 challenge on single image super-resolution: Dataset and study,'' in \emph{Proceedings of the IEEE conference on computer vision and pattern recognition workshops}, 2017, pp. 126--135.

\bibitem{nah2017deep}
S.~Nah, T.~Hyun~Kim, and K.~Mu~Lee, ``Deep multi-scale convolutional neural network for dynamic scene deblurring,'' in \emph{Proceedings of the IEEE conference on computer vision and pattern recognition}, 2017, pp. 3883--3891.

\bibitem{chang2004super}
H.~Chang, D.-Y. Yeung, and Y.~Xiong, ``Super-resolution through neighbor embedding,'' in \emph{Proceedings of the 2004 IEEE Computer Society Conference on Computer Vision and Pattern Recognition, 2004. CVPR 2004.}, vol.~1.\hskip 1em plus 0.5em minus 0.4em\relax IEEE, 2004, pp. I--I.

\bibitem{timofte2013anchored}
R.~Timofte, V.~De~Smet, and L.~Van~Gool, ``Anchored neighborhood regression for fast example-based super-resolution,'' in \emph{Proceedings of the IEEE international conference on computer vision}, 2013, pp. 1920--1927.

\bibitem{timofte2015a+}
------, ``A+: Adjusted anchored neighborhood regression for fast super-resolution,'' in \emph{Computer Vision--ACCV 2014: 12th Asian Conference on Computer Vision, Singapore, Singapore, November 1-5, 2014, Revised Selected Papers, Part IV 12}.\hskip 1em plus 0.5em minus 0.4em\relax Springer, 2015, pp. 111--126.

\bibitem{lim2017enhanced}
B.~Lim, S.~Son, H.~Kim, S.~Nah, and K.~Mu~Lee, ``Enhanced deep residual networks for single image super-resolution,'' in \emph{Proceedings of the IEEE conference on computer vision and pattern recognition workshops}, 2017, pp. 136--144.

\bibitem{martin2001database}
D.~Martin, C.~Fowlkes, D.~Tal, and J.~Malik, ``A database of human segmented natural images and its application to evaluating segmentation algorithms and measuring ecological statistics,'' in \emph{Proceedings eighth IEEE international conference on computer vision. ICCV 2001}, vol.~2.\hskip 1em plus 0.5em minus 0.4em\relax IEEE, 2001, pp. 416--423.

\bibitem{huang2015single}
J.-B. Huang, A.~Singh, and N.~Ahuja, ``Single image super-resolution from transformed self-exemplars,'' in \emph{Proceedings of the IEEE conference on computer vision and pattern recognition}, 2015, pp. 5197--5206.

\bibitem{matsui2017sketch}
Y.~Matsui, K.~Ito, Y.~Aramaki, A.~Fujimoto, T.~Ogawa, T.~Yamasaki, and K.~Aizawa, ``Sketch-based manga retrieval using manga109 dataset,'' \emph{Multimedia tools and applications}, vol.~76, pp. 21\,811--21\,838, 2017.

\bibitem{foi2007pointwise}
A.~Foi, V.~Katkovnik, and K.~Egiazarian, ``Pointwise shape-adaptive dct for high-quality denoising and deblocking of grayscale and color images,'' \emph{IEEE transactions on image processing}, vol.~16, no.~5, pp. 1395--1411, 2007.

\bibitem{sheikh2005live}
H.~Sheikh, ``Live image quality assessment database release 2,'' \emph{http://live. ece. utexas. edu/research/quality}, 2005.

\bibitem{xu2013unnatural}
L.~Xu, S.~Zheng, and J.~Jia, ``Unnatural l0 sparse representation for natural image deblurring,'' in \emph{Proceedings of the IEEE conference on computer vision and pattern recognition}, 2013, pp. 1107--1114.

\bibitem{hyun2014segmentation}
T.~Hyun~Kim and K.~Mu~Lee, ``Segmentation-free dynamic scene deblurring,'' in \emph{Proceedings of the IEEE conference on computer vision and pattern recognition}, 2014, pp. 2766--2773.

\bibitem{gong2017motion}
D.~Gong, J.~Yang, L.~Liu, Y.~Zhang, I.~Reid, C.~Shen, A.~Van Den~Hengel, and Q.~Shi, ``From motion blur to motion flow: A deep learning solution for removing heterogeneous motion blur,'' in \emph{Proceedings of the IEEE conference on computer vision and pattern recognition}, 2017, pp. 2319--2328.

\end{thebibliography}

\begin{IEEEbiography}
  [{\includegraphics[width=1in,clip,keepaspectratio]{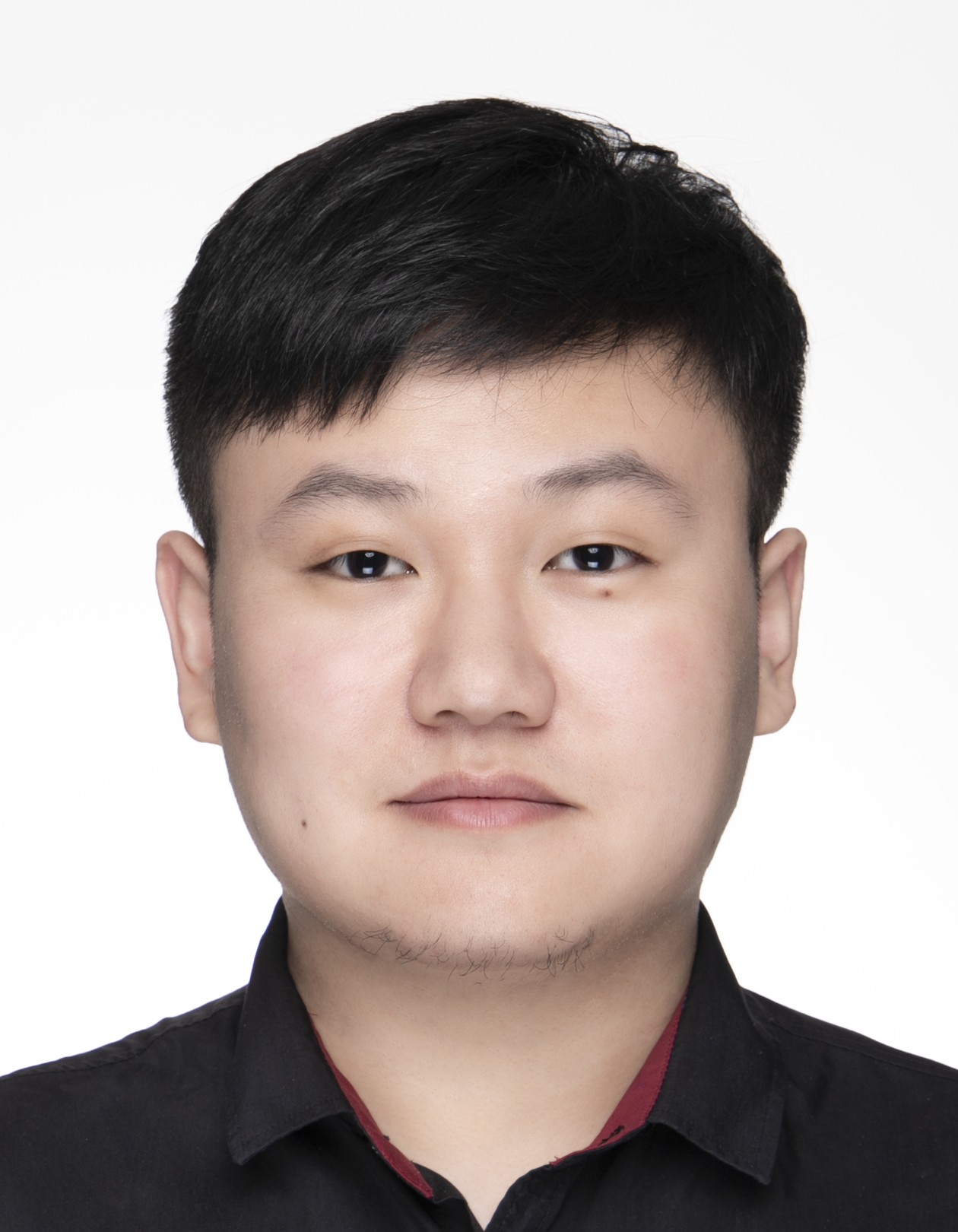}}]{Xi Zhang}
  (Member, IEEE) 
  received the B.Sc. degree 
  in mathematics and physics basic science 
  from the University of Electronic Science and Technology of China, in 2015, 
  and the Ph.D. degree 
  in electronic engineering 
  from Shanghai Jiao Tong University, China, in 2022.
  He was a postdoctoral fellow at McMaster University (Mac), Canada, from July 2022 to August 2024. 
  He is currently a Research Scientist with the Alibaba-NTU Global e-Sustainability CorpLab (ANGEL) 
  at Nanyang Technological University (NTU).
  His current research focuses on Green AI, particularly on efficient model design, sustainable system architectures, and resource-aware compression techniques.  
  He is also interested in broader challenges in deep learning, including domain generalization and visual reasoning.
  He has published more than 10 papers at the TOP Journals and conferences 
  such as T-PAMI, T-IP, NeurIPS, CVPR, etc. 
\end{IEEEbiography}

\begin{IEEEbiography}
  [{\includegraphics[width=1in,clip,keepaspectratio]{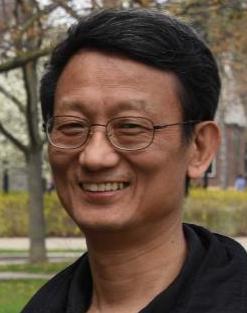}}]{Xiaolin Wu}
  (Life Fellow, IEEE) 
  received the B.Sc. degree in computer science 
  from Wuhan University, China, in 1982, and the Ph.D. degree in computer science 
  from the University of Calgary, Canada, in 1988. He started his academic career 
  in 1988. He was a Faculty Member with Western University, Canada, and New York 
  Polytechnic University (NYU-Poly). He is currently with McMaster University and 
  Southwest Jiaotong University. His research interests include image processing, 
  data compression, digital multimedia, low-level vision, and network-aware visual 
  communication. He has authored or co-authored more than 300 research articles and
  holds four patents in these fields. He served on technical committees for many 
  IEEE international conferences/workshops on image processing, multimedia, data 
  compression, and information theory. He was a past Associate Editor of IEEE 
  Transactions on Multimedia and IEEE Transactions on Image Processing.
\end{IEEEbiography}

\end{document}